\documentclass[sn-basic]{sn-jnl}

\jyear{2023}%

\theoremstyle{thmstyleone}%
\theoremstyle{thmstyletwo}%

\theoremstyle{thmstylethree}%

\usepackage{epsfig}
\usepackage{amsmath}

\usepackage{babel}
\usepackage{soul}

\usepackage{xcolor}

\begin{document}

\title[Patient Acceptance in Medical Robotic Imaging]{On the Importance of Patient Acceptance for Medical Robotic Imaging}


\author*[1]{\fnm{Christine} \sur{Eilers}}\email{christine.eilers@tum.de}
\equalcont{These authors contributed equally to this work.}

\author[1]{\fnm{Rob} \sur{van Kemenade}}\email{r.kemenade@tum.de}
\equalcont{These authors contributed equally to this work.}

\author[1]{\fnm{Benjamin} \sur{Busam}}\email{b.busam@tum.de}

\author[1]{\fnm{Nassir} \sur{Navab}}\email{nassir.navab@tum.de}

\affil[1]{\orgdiv{Chair for Computer Aided Medical Procedures and Augmented Reality}, \orgname{Technical University of Munich}, \orgaddress{\street{Boltzmannstr. 3}, \city{Garching near Munich}, \postcode{85748},\country{Germany}}}

\abstract{
\textbf{Purpose:} Mutual acceptance is required for any human-to-human interaction. Therefore, one would assume that this also holds for robot-patient interactions. However, the medical robotic imaging field lacks research in the area of acceptance. This work, therefore, aims at analyzing the influence of robot-patient interactions on acceptance in an exemplary medical robotic imaging system.

\textbf{Methods:} We designed an interactive human-robot scenario, including auditive and gestural cues, and compared this pipeline to a non-interactive scenario. Both scenarios were evaluated through a questionnaire to measure acceptance. Heart rate monitoring was also used to measure stress. The impact of the interaction was quantified in the use case of robotic ultrasound scanning of the neck. 

\textbf{Results:} We conducted the first user study on patient acceptance of robotic ultrasound. Results show that verbal interactions impacts trust more than gestural ones. Furthermore, through interaction, the robot is perceived to be friendlier. The heart rate data indicates that robot-patient interaction could reduce stress. 

\textbf{Conclusion:} Robot-patient interactions are crucial for improving acceptance in medical robotic imaging systems. While verbal interaction is most important, the preferred interaction type and content are participant-dependent. Heart rate values indicate that such interactions can also reduce stress. Overall, this initial work showed that interactions improve patient acceptance in medical robotic imaging, and other medical robot-patient systems can benefit from the design proposals to enhance acceptance in interactive scenarios.  
}


\maketitle

\section{Introduction}\label{sec_intro}
Acceptance is necessary for almost all human interactions. You, as the reader, accept that the authors provide honest and interesting information in this paper, while we, as the authors, accept that you will analyze the paper carefully and critically. Acceptance also extends to verbal and physical interactions. By accepting other's actions, you believe that these actions are reasonable and will not harm you. In a medical setting, you as the patient, would accept and trust the physician to perform an examination. This behavior entails that you accept clinical procedures. Furthermore, you accept that the physician has gathered enough knowledge to make a diagnosis, that he or she knows the correct diagnosis and treatment plan, and that he or she will not harm you with his or her actions. Acceptance is, therefore, a well-established concept in human-to-human interaction. 

With the increasing number of machines and robots entering everyday life, human-machine interactions also have to be analyzed. However, in contrast to accepting each other, humans do not inherently accept machines. In fact, acceptance of robots must be a prerequisite for successful human-robot interactions. Only if acceptance is achieved, new robotic solutions can be integrated into everyday life and interact with humans. 

Medical robotics has started to prove its advantages. Often, patient care could improve with these systems by increasing diagnostic accuracy and reducing the need for open surgeries. It is all the more astonishing that acceptance, especially focusing on patient acceptance, has not been explored widely in medical robotics. Even translational research, which focuses on moving research projects from the laboratory to their application environments, merely touches on this topic. Acceptance has been evaluated in robotic surgery~\cite{BenMessaoud, Knoop, McDermott} as well as in works which focus on the physician-robot relationship~\cite{Marin, Vichitkraivin}. Attia et al.~\cite{Attia} introduced a framework for trusted autonomy within surgical robotics. However, in this work, only the robot-surgeon relationship is analyzed. Torrent-Sellens et al.~\cite{Torrent-Sellens} analyzed trust in robot-assisted surgeries within Europe. Results show that introducing surgical systems mainly depends on the patient's wishes. However, in surgical robotic systems, patients are not interacting directly with the system. In robotic surgeries, in general, the patient is anesthesized and acceptance is only a matter of population trust in the statistical outcome of such robotic surgeries. Acceptance plays a different role when the robot interacts with a fully awake patient. Elderly care describes such a scenario and acceptance for robotic systems has been evaluated in this area~\cite{Hall, Broadbent, Wrobel, Koceski}. In a clinical context, Bodenhagen et al.~\cite{Bodenhagen, Fischer} analyzed the effect of different interaction channels in trusting the robot. It was shown that transparency can increase trust and transparency itself can be increased through communication. Weigelin et al.~\cite{Weigelin} concluded that vocal interactions increase trust compared to only kinesthetic interactions. 

Please note that different from usual scenarios, in cases that include fully awake patients, the robotic system wishes to interact with the patient, not the other way around. Therefore we are referring to these scenarios as robot-patient cases rather than patient-robot ones.

This paper presents an analysis of acceptance in medical robotic systems, focusing on systems that require robot-patient interaction. We evaluate the effect of interaction and communication on acceptance in the setting of a robotic ultrasound (US) scanning procedure. Furthermore, we analyze two different evaluation metrics. 

The research in robotic ultrasound currently strongly focuses on methodological improvements in scanning algorithms and data post-processing. Li et al.~\cite{Li} summarized the latest work in robotic US research. Research topics include the aspects of force~\cite{virga2016automatic, Jiang2021DeformationAwareR3}, compounding optimization~\cite{Sutedjo}, anatomy extraction~\cite{Zielke}, optimal view navigation~\cite{Bi}, sensor fusion~\cite{esposito2015cooperative, hennersperger2016towards, zettinig2015multimodal, esposito2016multimodal} and collaborative robotics~\cite{busam2015stereo}. So even though this field shows much progress, the topic of acceptance is not well studied. 

This work aims to close this gap and explores the topic of acceptance in medical robotic imaging. We exemplify this with a robotic ultrasound scanning procedure. However, insights can be transferred to similar robot-patient interaction scenarios. Specifically, we will look at acceptance in the context of a robotic ultrasound scanning of the neck. The neck is an especially critical and sensitive area. We, therefore, can assume that acceptance-enhancing aspects for this anatomy can be easily translated to other anatomies and examinations.

\section{Methods}\label{sec_methods}
In this work, we make the first attempt to objectively measure patient acceptance for medical robotic imaging systems. Through acceptance, a person believes that actions performed by the robot are useful and will not harm the person. Even though this definition is very understandable, it is difficult to quantify. The robot's actions can seem more apparent or unclear, while relying on something is very subjectiv, depending on previous experiences, and other factors. Therefore, analyzing the topic of acceptance is three-fold: How can we ensure acceptance in medical robotic systems? Which parameters in the interaction can be varied to influence acceptance? Moreover, how do we measure acceptance? In the following, we propose a pipeline and an associated user study to answer these questions.

\begin{figure}
    \centering
    \includegraphics[width=\textwidth]{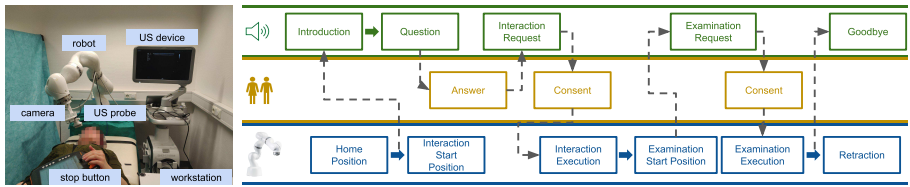}
    \caption{Left: the setup, in which data from the camera is used to run OpenPose to detect hand and neck keypoints. The camera and US probe are attached to the robotic manipulator through a custom-printed holder. A stop button is used as a safety measure to stop the robot's motion at any time. Right: Overview of the proposed method, divided into verbal cues (green, top row), robot actions (blue, bottom row), and human interaction (yellow, middle row). Dotted lines represent the interactions between the three main parts.}
    \label{fig:overview}
\end{figure}

The presented pipeline (Fig.~\ref{fig:overview}) allows for gestural and visual interaction, and communication with the robot. Therefore, these three types can be analyzed. Furthermore, the pipeline permits to tweak the content of each block separately. All interactions and communications are designed based on an analysis of sonographers' usual interactions with patients in routine clinical settings. Similar to a sonographer, the robot first introduces itself and asks whether the patient has already received a US thyroid scan and is familiar with the process. Taking inspiration from a human-to-human setting, we believe that this superficial conversation can relax the situation. The robot then waits for an answer before continuing with the interaction step. This communication gives the patient the feeling that the robot is taking the patient's response into account. Similar to a human sonographer, the robot executes an initial gestural interaction before examining the area of interest. In this case, the robot suggests a friendly high five to implicitly show action and location awareness while further connecting to the patient. Upon receiving consent from the patient, the robot executes the high five but does not touch the patient's hand due to hygiene regulations. After this interaction, the robot asks whether it is allowed to approach the neck for the US scan. Once again, the patient has to consent to this action while the robot shows awareness by reacting to the patient's reply. The robot then approaches the neck and performs the examination. Lastly, it retracts and says goodbye. 

The accompanying user study aims at comparing acceptance in interactive and non-interactive scenarios by evaluating an interactive and communicative, pipeline-based approach against the same task without any interaction. In the presented study, this task is defined as the movement of the robotic US probe to the neck to enable US scanning of the thyroid. In the non-interactive case, the robot simply approaches the neck, while the interactive version is executed as described above. 

Next to proposing an interactive and communicative method, both methods also have to be evaluated. For this, we propose using questionnaires and heart rate recordings. The questionnaire gathers the participants' opinions on both approaches. We are adding an additional measurement by incorporating a heart rate analysis. In the remaining manuscript, interaction also includes communication aspects if not mentioned otherwise.

\section{Experiments and Results}\label{sec_exp}

\subsection{Setup}\label{subsec_setup}
The setup can be seen in Figure~\ref{fig:overview}. It consists of a robotic manipulator (KUKA LBR iiwa, KUKA AG, Augsburg, Germany), a RGB-D camera (RealSense D435, Intel, USA), attached to the robotic manipulator by a custom-made holder, a ultrasound machine (Siemens ACUSON Juniper, Siemens Healthineers, Germany) with a linear transducer mounted to the robot and a workstation (Ubuntu 18.04, ROS melodic). The user is also provided with an emergency stop, which immediately stops the robot if needed. The workstation runs OpenPose~\cite{Cao,simon2017hand} and the robot navigation. OpenPose detects human body, facial and hand keypoints in real time. The robot is controlled via iiwa\_stack~\cite{hennersperger2017towards}. Furthermore, a safety margin of two centimeters was introduced for the points of contact on hand and neck. 

\subsection{Evaluation}\label{subsec_eval}
We evaluated the interactive method against a method without interaction, both in which the task finished with the robot moving towards the neck. Acceptance in the system was evaluated based on the measured heart rate and a questionnaire which is detailed in Table~\ref{tab:statements}. For the former, a heart rate monitor was attached to the participant's chest (HRM-Dual, Garmin, USA). The heart rates were compared during the movement of the US probe to the neck for both versions. The questionnaire was based on work by Schaefer et al.~\cite{Schaefer}, which focuses on long-term and collaborative interaction, but was adapted to our case. The original questionnaire consists of 40 questions with a ranking from $0 - 100 \%$ dependant on how often the robot fulfills the statement. Similar to their proposed 14-item subset, we chose 16 statements from the original questionnaire which were applicable to our setup. Thereby, we ,for example, removed statements that analyzed a collaborative interaction. Furthermore, we adapted the answers to a 5 point Likert scale (\ref{tab:likert}), because each participant experiences each procedure only once.

\begin{table}[h]
    \centering
    \begin{tabular}{l l}
    \hline
         1. The robot is responsible. & 9. The robot acts consistently. \\
         2. The robot is incompetent. & 10. The robot functions successfully. \\
         3. The robot is friendly. & 11. The robot malfunctions. \\
         4. The robot is reliable. & 12. The robot communicates clearly. \\
         5. The robot is pleasant. & 13. The robot meets the needs of the task. \\
         6. The robot is unresponsive. & 14. The robot provides appropriate information. \\
         7. The robot is autonomous. & 15. The robot communicates with people. \\
         8. The robot is predictable. & 16. The robot performs exactly as instructed. \\
    \hline
    \end{tabular}
    \caption{Questionnaire statements, adapted from~\cite{Schaefer}. The users were asked to rank the statements in five categories from strongly agree to strongly disagree.}
    \label{tab:statements}
\end{table}

\begin{table}[]
    \centering
    \begin{tabular}{|l|c|c|c|c|c|}
        \hline
        questionnaire & strongly disagree & disagree & neutral & agree & strongly agree \\
        \hline
        value & 1 & 2 & 3 & 4 & 5 \\
        \hline
    \end{tabular}
    \caption{Likert scale answer and assigned trust score value.}
    \label{tab:likert}
\end{table}

The trust score was computed by assigning a value to each option of the Likert scale (Table~\ref{tab:likert}). The results of Statements 2, 6, and 11 are inverted, respectively. The term 'pleasant' could be interpreted subjectively by each participant. Statement 10 and 11 were both included to enable a differentiation between malfunctioning and functioning in a way that creates bigger expectations than normal functioning. The overall trust score is then computed as an average of all statements.

For the user study, 20 participants were recruited (10 male, 10 female, 20-33 years, mean: 25.8 ($\pm 2.62$) years), resulting in an interdisciplinary group of medical and computer science researchers. All of them were unfamiliar with the system. The experience with robotic applications ranged from low (14 participants) to medium (4 participants) to high (2 participants). All participants experienced both approaches in a randomized order, directly one after the other. Participants were informed about the goal and process of the study. No preliminary training was necessary.

\subsection{Results}\label{subsec_results}
\begin{figure}[h]
    \centering
    \includegraphics[scale=0.50]{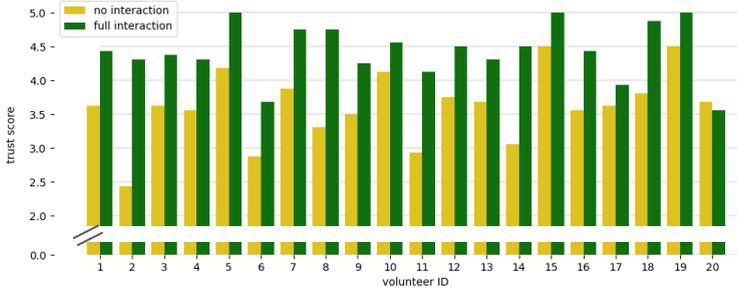}
    \caption{Block diagram showing the overall trust score for each participant in the no interaction (yellow) and full interaction (green) cases.}
    \label{fig:que_res}
\end{figure}

\begin{figure}[h]
    \centering
    \includegraphics[scale=0.35]{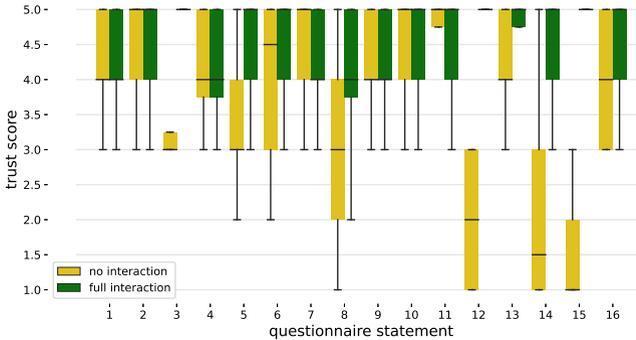}
    \caption{Box diagram showing the results for each questionnaire statement for the no interaction (yellow) and full interaction case (green).}
    \label{fig:que_box}
\end{figure}

The average trust score over all participants amounted to $3.61 \pm 0.50$ and $4.43 \pm 0.40$ for the no interaction and the full interaction case, respectively. Figure~\ref{fig:que_res} shows the overall trust score of each participant for no interaction (yellow) and the full interaction (green) scenario. In 19 out of 20 cases, the complete interaction case led to a higher trust score. Interestingly, robotic experience and gender parameters do not correlate with the trust score. Furthermore, we did not see a significant training improvement from the first to the second testing approach in any order. Figure~\ref{fig:que_box} shows trust scores dependent on each questionnaire statement. It can be seen that statements 3, 8, 12, 14, and 15 show the most significant increase in trust score in the interactive scenario compared to no interaction.

\begin{figure}[h]
    \centering
    \includegraphics[scale=0.35]{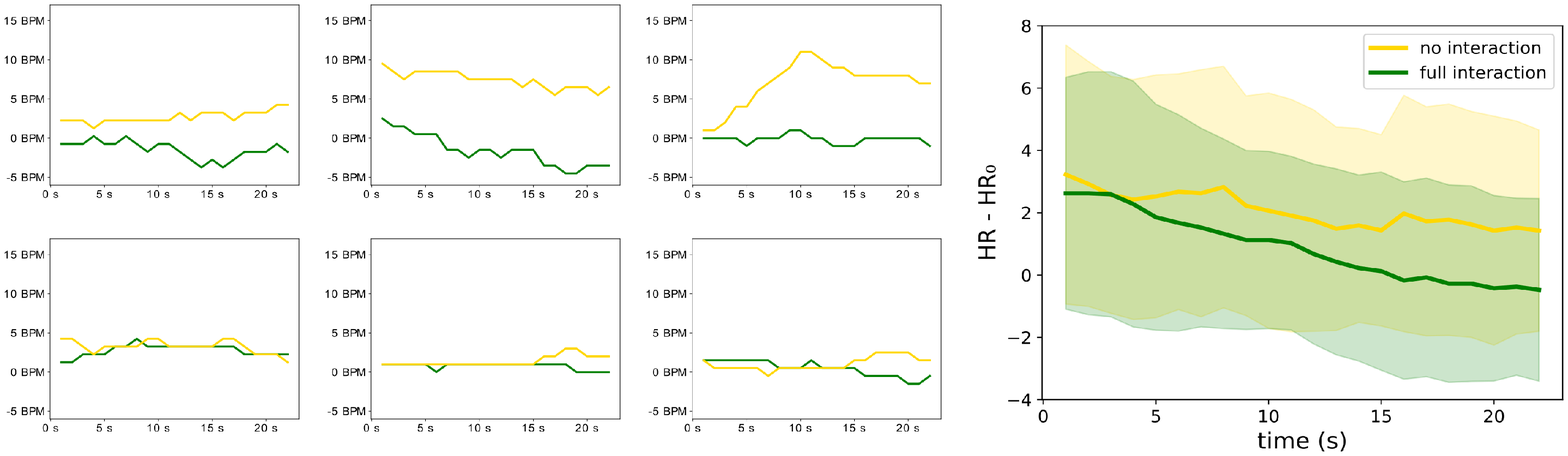}
    \caption{Comparison between the resting heart rate ($HR_0$) and the heart rate during the procedure ($HR$). Right: Mean and standard deviation of the heart rate difference to the resting state over all participants. Left: examples from individual participants, top row: the full interaction is favorable, bottom row: both options are favorable.}
    \label{fig:hr}
\end{figure}

To objectively evaluate the stress level of the participants, the heart rate values are analyzed. First, the median in a resting state was computed for each participant. Then the difference between median and heart rates during the procedure was computed. Fig.~\ref{fig:hr} shows the average result of this comparison over all participants and some examples for specific participants. The average heart rate difference compared to rest state shows a decrease in heart rate of 1.8 BPM in the no interaction condition and a stronger decrease of 3.1 BPM in the full interaction condition.

\section{Discussion}\label{sec_disc}
Performing and evaluating the user study resulted in different learnings. 

Results from the questionnaire are promising, showing that human-kind interactions with the robotic system increase the trust score. By analyzing the questionnaire statements separately, it is apparent that statements 12, 14, and 15 focus on communication. Therefore, the results show that the interactive method entails better communication with the participants. Statement 8 concerns the predictability of the system. The increased trust score shows that communicating information is beneficial. Lastly, statement 3 concerns the friendliness of the system and therefore touches on the aspect of acceptance on an emotional level. The results show that with the proposed interaction, the acceptance aspect can also be increased. In three statements (7, 10, and 11), the no interaction scenario achieves a marginally higher trust score than the full interaction. The results of statements 10 and 11 can be explained by two questionnaire responses that could be considered as outliers. In these cases, the robot's motion did not perform typically, thereby decreasing the trust score in the robot to function successfully. Statement 7 evaluates how autonomous the robot is being perceived. The results show that by introducing interactions, the robotic system is perceived to be less autonomous.

Even though the average trust scores for the full interaction of statements 5, 8, 9, 12, and 16 are high ($4.3, 4.1, 4.25, 4.75, 4.45$, respectively), all statements contain responses with a trust score of two. These outliers indicate that the content and type of different interactions are perceived subjectively. Additionally, participant feedback suggests that communication was more appreciated than the initial gestural high five. 

To objectively evaluate the impact of interactions on patient acceptance, we also gathered heart rate data, specifically analyzing the difference in heart rates compared to resting heart rates. The results show a slightly stronger decrease in heart rates in the full interaction scenarios. This decrease could marginally indicate that interaction reduces the stress level during the procedure. However, most individual results only show a slight difference in heart rates (see Fig.~\ref{fig:hr}, left). Therefore, choosing a more stress-reducing scenario per participant is difficult. By including a heart rate analysis, we showed a different measurement to the questionnaire. The former analysis marginally supports the questionnaire results, as a reduction in stress can also be related to a higher acceptance.

This study shows that interactions can improve the acceptance of a robotic ultrasound scanning system. By choosing the neck as the area of interest, this application tackles a susceptible and critical structure of the human body. Therefore, we believe that this study's results can also be translated to robotic ultrasound scanning of other anatomies. Furthermore, we believe that insights from this study can be used to improve the acceptance of different robotic imaging systems in general. In the future, more extensive and diverse user studies are necessary to evaluate the interaction effect on different populations and analyze the effect of different interaction and communication types. 

Based on participants' feedback, the robot trajectory also influences the acceptance of the robotic system. In future works, this aspect should therefore be analyzed. A more human-mimicking robot motion could, for example, increase acceptance. 

This initial study was conducted with young volunteers and showed promising results on the impact of interaction and communication on acceptance. However, this cohort does not resemble most actual patient cohorts in thyroid diagnostics. In general, these patients are older and less used to modern technology. Future studies should therefore be performed with patient groups to analyze the effect of interactions and communication in more detail and in a more realistic environment. Insights from these studies would then also help to introduce more patient- and examination-specific interactions. A bigger cohort size would also allow for different ablation studies, for example, analyzing the effect of communication to no communication, the effect of different voices or the effect of different verbal cues.

Lastly, this paper proposed an initial robot-patient interaction to increase acceptance in robotic imaging systems. However, the results showed that the type and content of interactions could be perceived differently. Therefore, future interactions could be tailored to each specific patient. In that case, previous interactions with robotic systems and other factors can be used to define a patient- and examination-specific interaction. Furthermore, different patient information, such as name and age, could be extracted from patient files to enable more realistic and pleasant communications. This additional information would allow researchers to extend the current sequential workflow to a branched one that allows for deviations in answers and interactions. So do not be surprised, if the robotic sonographer asks you:'Do you feel better since we last met in October?'

\section{Conclusion}\label{sec_concl}
The presented work analyzed the impact of robot-patient interaction on acceptance in robotic ultrasound scanning. A questionnaire was used to evaluate acceptance, and heart rate monitoring was used for stress evaluation. Results show that interactions before the examination can improve patient acceptance. Communication is crucial in increasing acceptance, and the robotic system is perceived as friendlier if patient interactions are taken into account. The heart rate analysis shows that interactions could reduce the patient's stress level. Learnings from this work can be used to adapt other medical robotic systems requiring patient interaction and pave the path towards patient- and examination-specific interaction designs for these systems. 

\backmatter

\bmhead{Supplementary information}
A supplementary video file is submitted.  

\bmhead{Acknowledgments}

The authors would like to thank all volunteers for participating in the study.

\section*{Declarations}
The authors declare no conflict of interest. \\
Consent for participation and publication of the anonymized data was taken from all participants. No ethical approval was required for this user study.

\bibliographystyle{unsrt}
\bibliography{references}

\end{document}